\documentclass[runningheads]{llncs}
\usepackage[normalem]{ulem}
\usepackage[T1]{fontenc}
\usepackage[nocompress]{cite}
\usepackage{amsfonts}
\usepackage{multicol}
\usepackage{multirow}
\usepackage{booktabs}
\usepackage{array}
\usepackage{arydshln}
\usepackage{xcolor}
\usepackage{colortbl}
\usepackage{algorithmicx}
\usepackage{subcaption}
\usepackage{algorithm}
\usepackage[noend]{algpseudocode}
\usepackage{amsmath}
\usepackage{amssymb}
\usepackage{tcolorbox}

\usepackage{graphicx,verbatim}
\usepackage[misc]{ifsym}
\renewcommand{\arraystretch}{0.9}
\newcommand{\cmark}{\ensuremath{\checkmark}}
\newcommand{\xmark}{\ensuremath{\times}}

\begin{document}
\title{ICGF-CXRMate: Image-Conditioned Clinical Context Compression and Gated Fusion for Chest X-ray Report Generation}
\titlerunning{ICGF-CXRMate}

\author{Vu Minh Tran$^{1,2,*}$, Thuc Huynh$^{1,2,*}$, Son Thanh Luu$^{1,2}$, Doanh C. Bui$^{1,2}$}
\authorrunning{Tran et al.}
\institute{$^1$University of Information Technology \\ $^2$Viet Nam National University Ho Chi Minh City \\
$^*$ These authors contributed equally to this work
}


\maketitle

\begin{abstract}
Auxiliary patient data can improve chest X-ray report generation, but directly concatenating heterogeneous clinical information with visual tokens produces long, variable-length decoder prompts and may expose the model to redundant or image-irrelevant context. We propose ICGF-CXRMate, an image-conditioned clinical context compression and gated fusion framework. Learnable queries are first conditioned on the current chest X-ray representations and then attend to the patient-data sequence, compressing it into a fixed number of clinical context tokens. Gated cross-attention selectively integrates these tokens into the visual stream, while the compressed tokens are retained in the decoder prompt as a direct clinical pathway. Experiments on linked MIMIC-CXR and MIMIC-IV-ED examinations show that ICGF-CXRMate improves report quality while reducing the average prompt length. In Stage~2, it improves RadGraph-F1 from 24.00 to 28.86, CXR-BERT similarity from 56.86 to 62.80, and BLEU-4 from 5.14 to 7.96 compared with CXRMate-ED, while reducing the average prompt length from 373.89 to 288.37 tokens, corresponding to a 22.9\% reduction. After self-critical sequence training, ICGF-CXRMate outperforms CXRMate-ED on six of eight report-quality metrics in Stage~3. Ablation studies confirm the complementary contributions of image-conditioned compression, gated fusion, and direct retention of the compressed clinical tokens.
\keywords{Radiology report generation \and Chest X-ray \and Multimodal
learning \and Clinical context compression}
\end{abstract}

\section{Introduction}
\label{sec:introduction}

Chest X-ray (CXR) report generation aims to automatically produce a
radiology report from one or more images in a CXR examination. Such systems
may assist radiologists and improve reporting efficiency amid increasing
imaging workloads~\cite{jones2021chest}. The field has progressed from
conventional image-to-text models toward approaches that jointly process
multiple radiographs, optimize clinical consistency, and employ
radiology-specific large vision-language models
~\cite{miura2021improving,delbrouck2022improving,pellegrini2023radialog,hyland2023maira1,bannur2024maira2}.

Recent studies further incorporate information beyond the current
radiographs. Longitudinal methods use prior examinations to model interval
changes and disease progression
~\cite{nicolson2024longitudinal,hou2023recap,wang2025hergen,zhang2025libra}, while other approaches use the examination indication to
guide generation toward patient-specific clinical concerns
~\cite{nguyen2023pragmatic,liu2024sei}. These findings show that contextual
information can improve the clinical relevance of generated reports.

In clinical practice, radiologists interpret images together with presenting
symptoms, examination indication, clinical history, and medication
records~\cite{castillo2021effect}. CXRMate-ED~\cite{nicolson2025impact}
introduced a unified representation for heterogeneous patient data and
identified triage, reconciled medicines, indication, and clinical history
as effective inputs for CXR report generation. Following this evidence, we
adopt the same four sources and focus on improving how their variable-length
representations are compressed and integrated with visual tokens.

CXRMate-ED directly concatenates the complete patient-data sequence with the
visual tokens, producing a decoder prompt whose length grows with the
available clinical information. Larger inputs may degrade performance due
to prompt growth, attention dilution, and potentially confounding auxiliary
evidence~\cite{nicolson2025impact,qin2022devil}. This motivates a mechanism
that selects image-relevant clinical context and controls its influence on
the visual representations.

To address these limitations, we propose \textbf{ICGF-CXRMate}, an
image-conditioned clinical context compression and gated fusion framework.
Learnable queries are conditioned on the current CXR representations and
then retrieve relevant information from the patient sequence, producing a
fixed number of clinical context tokens. Gated cross-attention selectively
integrates this context into the visual stream, while the compressed tokens
are retained as a direct clinical pathway to the decoder.

Experiments on linked MIMIC-CXR and MIMIC-IV-ED examinations show that
ICGF-CXRMate reduces the average prompt length from 373.89 to 288.37 tokens,
corresponding to a 22.9\% reduction. In Stage~2, it improves RadGraph-F1
from 24.00 to 28.86, CXR-BERT similarity from 56.86 to 62.80, and BLEU-4
from 5.14 to 7.96 over CXRMate-ED. After self-critical sequence training,
it outperforms CXRMate-ED on six of eight report-quality metrics in
Stage~3.

\noindent\textbf{The main contributions are as follows:}
\begin{itemize}
    \item We propose an image-conditioned clinical context compression
    module that converts variable-length auxiliary patient data into a
    fixed number of radiographically relevant clinical tokens.

    \item We introduce gated cross-attention fusion to selectively integrate
    the compressed context into the visual stream while preserving a direct
    clinical-token pathway to the decoder.

    \item Experiments and ablation studies demonstrate improved report
    quality, a 22.9\% reduction in average prompt length, and consistent
    benefits after self-critical sequence training.
\end{itemize}

\section{Related Work}
\label{sec:related_work}

\noindent\textbf{Chest X-ray Report Generation.}
CXR report generation has progressed from conventional image
encoder--language decoder models toward clinically informed multimodal
architectures. Semantic rewards based on clinical entities and relations
improve factual completeness and consistency
~\cite{miura2021improving,delbrouck2022improving}, while entropy-regularized
self-critical sequence training reduces overfitting to common report
expressions~\cite{nicolson2024rrg24}. Recent approaches increasingly employ large language models,
radiology-specific visual representations, and external knowledge.
RaDialog supports report generation and interactive assistance, MAIRA-1 and
MAIRA-2 adapt large multimodal models to CXR reporting and spatial grounding,
and KARGEN incorporates disease-related knowledge graphs
~\cite{pellegrini2023radialog,hyland2023maira1,bannur2024maira2,li2024kargen}. These methods improve visual understanding and clinical
grounding, but primarily focus on the current radiographs.

\noindent\textbf{Longitudinal and Clinical Context.}
Longitudinal methods incorporate previous examinations to capture interval
changes and disease progression. CXRMate processes prior images and reports,
RECAP models observation progression across consecutive radiographs, and
HERGen and Libra integrate historical information through visit-level or
temporal alignment mechanisms
~\cite{nicolson2024longitudinal,hou2023recap,wang2025hergen,zhang2025libra}. Clinical text can also guide generation toward the purpose of the current
examination. Pragmatic report generation uses the indication section, while
SEI combines CXR images, similar historical cases, and patient-specific
indications~\cite{nguyen2023pragmatic,liu2024sei}. CXRMate-ED further
represents numerical, categorical, textual, temporal, and image data in a
unified multimodal prompt~\cite{nicolson2025impact}. However, it provides
the complete patient-data sequence directly to the decoder, making prompt
length dependent on the available records and potentially introducing
confounding evidence. In contrast, our work compresses this sequence and
controls its interaction with the visual representations.

\noindent\textbf{Context Compression and Multimodal Fusion.}
Prompt-compression methods reduce input length while preserving information
required for generation. LLMLingua performs coarse-to-fine token-level
compression for long textual prompts \cite{jiang2023llmlingua}, but does not
explicitly model whether clinical context is relevant to the current
radiographic evidence. Multimodal architectures also use learnable queries and latent tokens to
produce compact representations. Flamingo transforms variable-sized visual
inputs into a fixed number of tokens using a Perceiver Resampler and gated
cross-attention~\cite{alayrac2022flamingo}, while BLIP-2 uses a Querying
Transformer to bridge a frozen image encoder and language model
~\cite{li2023blip2}.

ICGF-CXRMate differs in both the compressed modality and the conditioning
signal. Rather than compressing visual representations or text-only prompts,
it conditions learnable queries on the current CXR representations and uses
them to compress heterogeneous patient data. The resulting clinical tokens
are selectively fused into the visual stream while remaining directly
accessible to the report decoder.

\section{Methodology}
\label{sec:methodology}

\begin{figure}[t]
    \centering
    \includegraphics[width=\textwidth]{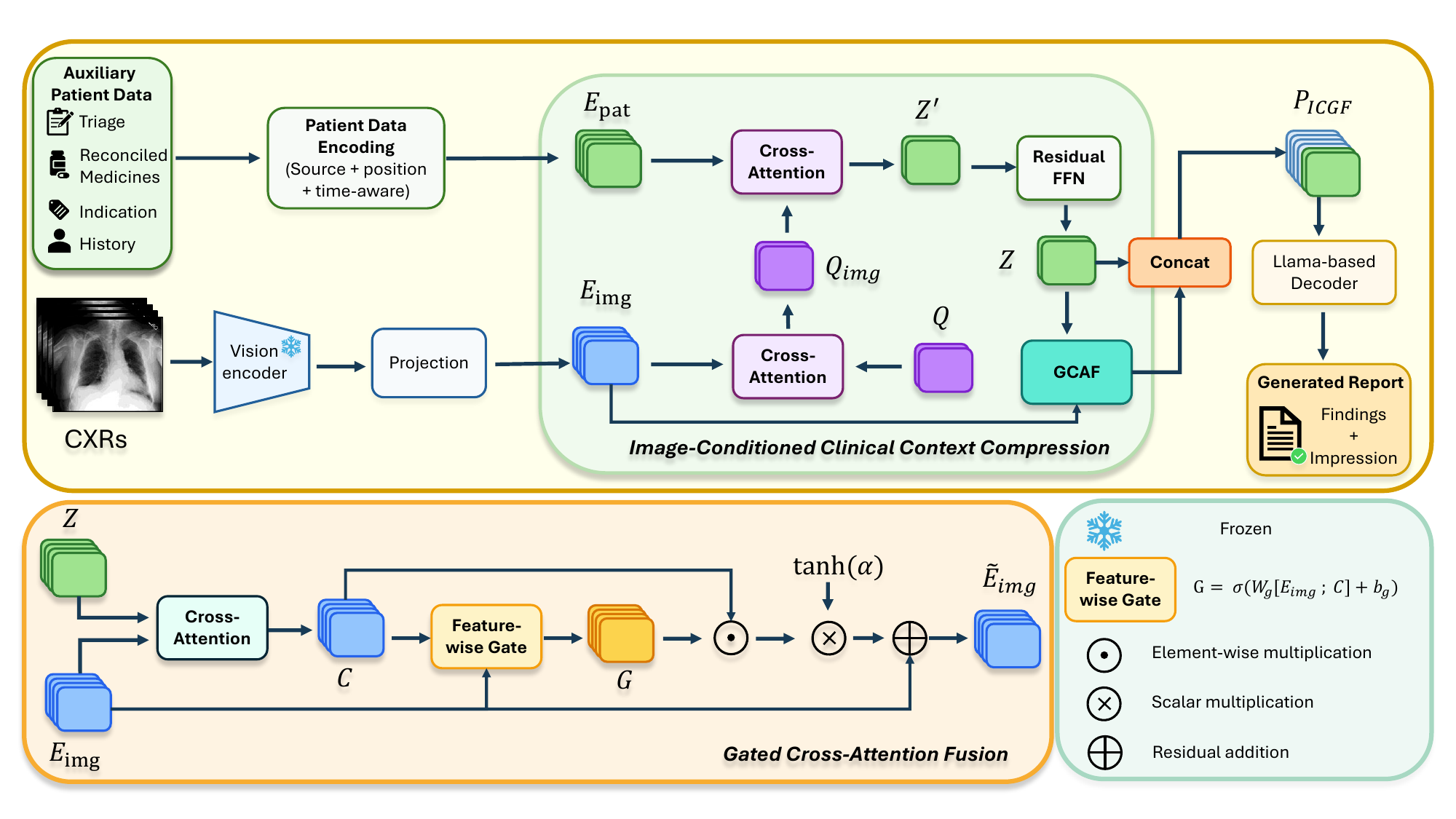}
    \caption{\textbf{Overview of ICGF-CXRMate.}
The model encodes CXR images and heterogeneous auxiliary patient data into
visual and clinical representations. Image-conditioned clinical context
compression (Sec.~\ref{sec:image_conditioned_compression}) uses
image-adapted queries to summarize the variable-length patient sequence into
a fixed set of clinical tokens $Z$. Gated cross-attention fusion
(Sec.~\ref{sec:gated_fusion}) selectively integrates the compressed context
into the visual representations to obtain
$\widetilde{E}_{\mathrm{img}}$. The resulting multimodal prompt
$P_{\mathrm{ICGF}}=[Z;\widetilde{E}_{\mathrm{img}}]$ is passed to the
Llama-based decoder for report generation.}
    
    \label{fig:main_method}
\end{figure}

We present ICGF-CXRMate, an extension of CXRMate-ED
for chest X-ray report generation using auxiliary patient data.
As illustrated in Fig.~\ref{fig:main_method}, the proposed method first
uses radiographic evidence to guide the compression of heterogeneous
clinical information and subsequently integrates the compressed context
into the visual representations through gated cross-attention. This design
reduces the patient-data portion of the decoder prompt while preserving
direct access to both visual evidence and compact clinical context.

\subsection{Baseline: CXRMate-ED}
\label{sec:baseline}

CXRMate-ED~\cite{nicolson2025impact} generates radiology reports from
chest radiographs and heterogeneous auxiliary patient data. Given a CXR
study $\mathcal{I}=\{I_s\}_{s=1}^{S}$ containing $S$ images, a vision
encoder followed by a layer-normalized projection produces the visual
token sequence

\begin{equation}
E_{\mathrm{img}}
=
\left[
\pi\!\left(f_{\mathrm{img}}(I_1)\right);
\ldots;
\pi\!\left(f_{\mathrm{img}}(I_S)\right)
\right]
\in
\mathbb{R}^{L_{\mathrm{img}}\times H},
\label{eq:image_embedding}
\end{equation}

where $L_{\mathrm{img}}$ is the total number of image tokens and $H$ is
the decoder hidden dimension. Auxiliary information, including triage,
reconciled medicines, indication, and clinical history, is encoded as a
variable-length patient sequence
$E_{\mathrm{pat}}\in\mathbb{R}^{L_{\mathrm{pat}}\times H}$.
Numerical and categorical variables are represented using source-specific
feed-forward encoders, whereas textual variables use the decoder token
embeddings. Source-type and temporal embeddings are incorporated to
identify the origin and temporal relevance of each clinical input.

CXRMate-ED directly concatenates the complete patient sequence with the
visual tokens to form the decoder prompt:

\begin{equation}
P_{\mathrm{base}}
=
\left[
E_{\mathrm{pat}};
E_{\mathrm{img}}
\right].
\label{eq:baseline_prompt}
\end{equation}

Consequently, the prompt contains
$L_{\mathrm{pat}}+L_{\mathrm{img}}$ tokens and grows with the amount of
available patient information. This variable-length formulation increases
the decoder input size and may expose the model to redundant or
image-irrelevant clinical information.

\subsection{Image-Conditioned Clinical Context Compression}
\label{sec:image_conditioned_compression}

To obtain a compact patient representation that is relevant to the current
radiographic evidence, we condition a fixed set of learnable queries on the
image embeddings before retrieving information from the patient sequence.


Let $Q\in\mathbb{R}^{K\times H}$ denote a set of $K$ learnable clinical
queries. The queries first attend to the source-aware visual representations:

\begin{equation}
Q_{\mathrm{img}}
=
\operatorname{LN}
\left(
Q
+
\operatorname{MHA}
\left(
\operatorname{LN}(Q),
E_{\mathrm{img}}^{q},
E_{\mathrm{img}}^{q}
\right)
\right),
\label{eq:image_conditioned_query}
\end{equation}

where $\operatorname{MHA}(\cdot)$ denotes multi-head cross-attention,
$\operatorname{LN}(\cdot)$ denotes layer normalization, and
$E_{\mathrm{img}}^{q}$ denotes the visual tokens augmented with image
token-type embeddings for cross-attention, while the original
$E_{\mathrm{img}}$ is retained as the residual visual stream. This operation
adapts the initially generic queries to the radiographic content of the
current study.

The image-conditioned queries subsequently retrieve relevant information
from the variable-length patient sequence:

\begin{equation}
Z^{\prime}
=
\operatorname{LN}
\left(
Q_{\mathrm{img}}
+
\operatorname{MHA}
\left(
\operatorname{LN}(Q_{\mathrm{img}}),
E_{\mathrm{pat}},
E_{\mathrm{pat}}
\right)
\right).
\label{eq:patient_cross_attention}
\end{equation}

The retrieved representations are further refined by a residual feed-forward
block:

\begin{equation}
Z
=
\operatorname{LN}
\left(
Z^{\prime}
+
\operatorname{FFN}
\left(
\operatorname{LN}(Z^{\prime})
\right)
\right),
\qquad
Z
\in
\mathbb{R}^{K\times H},
\label{eq:compressed_context}
\end{equation}

where $\operatorname{FFN}(\cdot)$ is a two-layer feed-forward network with
SiLU activation. The resulting sequence $Z$ contains a fixed number of
clinical context tokens, regardless of the original patient-data length
$L_{\mathrm{pat}}$. Attention masks are applied during both cross-attention
operations so that padded or unavailable inputs do not contribute to the
compressed representations.

Unlike direct query-based compression, the proposed module explicitly
conditions the retrieval of patient information on the current radiographic
evidence. Consequently, the same clinical source can be represented
differently depending on the visual findings of the corresponding study.

\subsection{Gated Cross-Attention Fusion}
\label{sec:gated_fusion}

To control how auxiliary clinical information influences the visual
representations, we introduce a gated cross-attention module that allows
each image token to selectively retrieve information from the compressed
clinical context $Z$.

The clinical context corresponding to each visual token is first computed as

\begin{equation}
C
=
\operatorname{MHA}
\left(
\operatorname{LN}(E_{\mathrm{img}}^{q}),
\operatorname{LN}(Z),
\operatorname{LN}(Z)
\right),
\qquad
C
\in
\mathbb{R}^{L_{\mathrm{img}}\times H}.
\label{eq:fusion_context}
\end{equation}

A feature-wise gate then determines how much of the retrieved clinical
context should be introduced into each dimension of the visual
representation:

\begin{equation}
G
=
\sigma
\left(
W_g
\left[
E_{\mathrm{img}};
C
\right]
+
b_g
\right),
\label{eq:fusion_gate}
\end{equation}

where $\sigma(\cdot)$ denotes the sigmoid function,
$[\cdot;\cdot]$ denotes concatenation along the feature dimension, and
$G\in\mathbb{R}^{L_{\mathrm{img}}\times H}$.

The fused visual representation is defined as

\begin{equation}
\widetilde{E}_{\mathrm{img}}
=
E_{\mathrm{img}}
+
\tanh(\alpha)
\left(
G\odot C
\right),
\label{eq:gated_fusion}
\end{equation}

where $\odot$ denotes element-wise multiplication and $\alpha$ is a
learnable scalar residual coefficient. The coefficient is initialized as
$\alpha=0$, such that $\tanh(\alpha)=0$ and

\begin{equation}
\widetilde{E}_{\mathrm{img}}
=
E_{\mathrm{img}}
\end{equation}

at initialization. The fusion module therefore behaves as an exact identity
mapping at the beginning of Stage 2 training and gradually learns how
strongly the clinical context should modify the pretrained visual
representations.

The final multimodal prompt retains both the compressed clinical context
and the fused visual tokens:

\begin{equation}
P_{\mathrm{ICGF}}
=
\left[
Z;
\widetilde{E}_{\mathrm{img}}
\right].
\label{eq:final_prompt}
\end{equation}

The resulting prompt contains $K+L_{\mathrm{img}}$ tokens, compared with
$L_{\mathrm{pat}}+L_{\mathrm{img}}$ tokens in CXRMate-ED. Retaining $Z$
provides the decoder with direct access to the compact patient context,
while $\widetilde{E}_{\mathrm{img}}$ provides visual features selectively
enhanced by the same clinical information.

A Llama-based decoder autoregressively generates the
\textit{Findings} and \textit{Impression} sections as a single sequence
separated by a dedicated section token. The multimodal prompt uses
non-causal attention, whereas causal masking is applied to the generated
report tokens.

\subsection{Training Objectives}
\label{sec:training_objectives}

The model is trained in three stages. In Stage 1, the vision encoder,
image-projection module, and language decoder are trained using CXR images
only. In Stage 2, the proposed model is initialized from the best Stage 1
checkpoint and extended with the patient-data encoders, image-conditioned
compressor, and gated fusion module. The UniFormer backbone is frozen during
Stage 2, while the remaining modules are optimized.

Stages 1 and 2 use teacher forcing with token-level negative log-likelihood:

\begin{equation}
\mathcal{L}_{\mathrm{TF}}
=
-
\sum_{t=1}^{T}
\log
p_{\theta}
\left(
y_t
\mid
y_{<t},
P
\right),
\label{eq:teacher_forcing_loss}
\end{equation}

where $Y=(y_1,\ldots,y_T)$ is the reference report,
$P=E_{\mathrm{img}}$ in Stage 1, and
$P=P_{\mathrm{ICGF}}$ in Stage 2.

\noindent\textbf{In Stage 3}, the model is initialized from the Stage 2
checkpoint and further optimized using self-critical sequence training
to improve report quality. The model generates a sampled report
$\widehat{Y}^{s}$, while a greedily decoded report
$\widehat{Y}^{g}$ is used as the baseline. The reinforcement-learning
objective is defined as

\begin{equation}
\mathcal{L}_{\mathrm{RL}}
=
-
\left[
r\!\left(\widehat{Y}^{s},Y\right)
-
r\!\left(\widehat{Y}^{g},Y\right)
\right]
\sum_{t=1}^{\widehat{T}}
\log
p_{\theta}
\left(
\widehat{y}^{s}_t
\mid
\widehat{y}^{s}_{<t},
P_{\mathrm{ICGF}}
\right),
\label{eq:reinforcement_loss}
\end{equation}

where $\widehat{T}$ is the length of the sampled report and
$r(\cdot)$ denotes the report-level reward. Since the
\textit{Findings} and \textit{Impression} sections have different
reporting functions, the reward is computed separately for the two
sections and combined using section-specific weights:

\begin{equation}
r\!\left(\widehat{Y},Y\right)
=
\alpha_f
r_f\!\left(\widehat{Y}_f,Y_f\right)
+
\alpha_i
r_i\!\left(\widehat{Y}_i,Y_i\right),
\label{eq:section_reward}
\end{equation}

where subscripts $f$ and $i$ denote the \textit{Findings} and
\textit{Impression} sections, respectively. For each section
$j\in\{f,i\}$, the composite reward is defined as

\begin{equation}
r_j\!\left(\widehat{Y}_j,Y_j\right)
=
\lambda_1
\operatorname{CB}\!\left(\widehat{Y}_j,Y_j\right)
+
\lambda_2
\operatorname{BS}\!\left(\widehat{Y}_j,Y_j\right)
+
\lambda_3
\operatorname{ARN}\!\left(\widehat{Y}_j\right),
\label{eq:composite_reward}
\end{equation}

where $\operatorname{CB}$, $\operatorname{BS}$, and
$\operatorname{ARN}$ denote CXR-BERT similarity, BERTScore, and the
absence of repeated n-grams, respectively.

\section{Experiments}
\subsection{Datasets}
\begin{figure}[t]
    \centering
    \includegraphics[width=\linewidth]{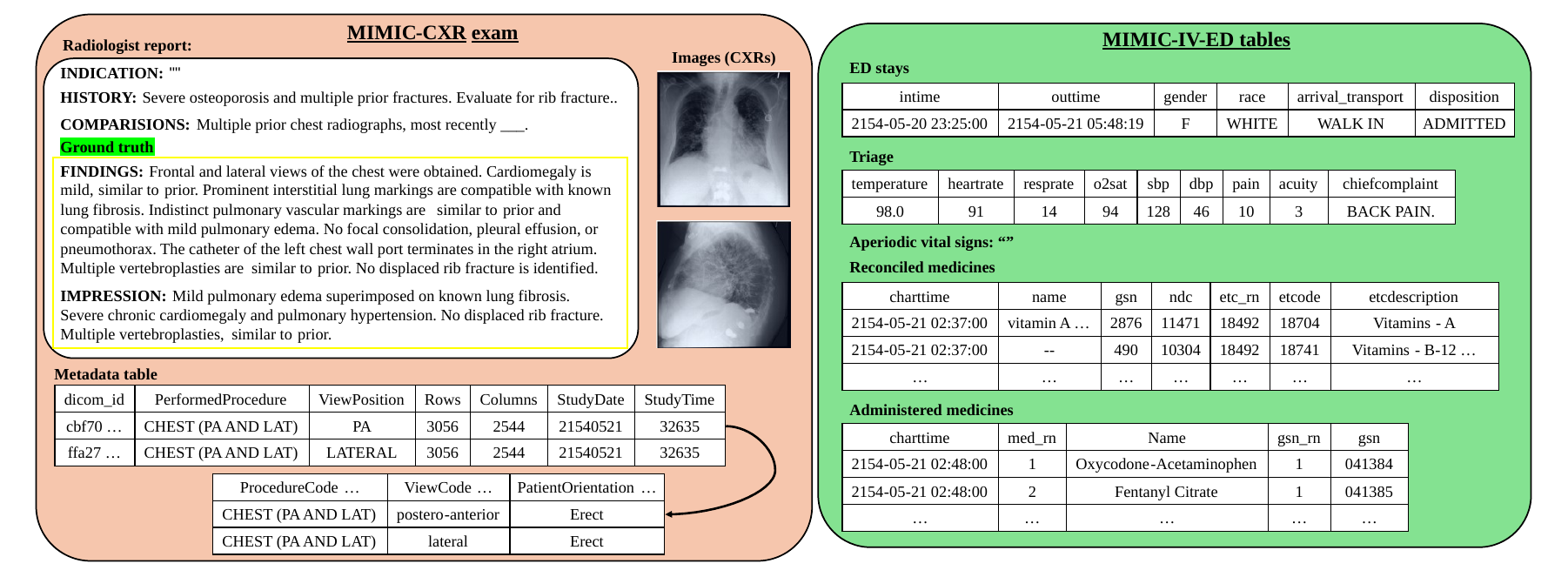}
    \caption{\textbf{Example of a linked MIMIC-CXR examination and
    MIMIC-IV-ED stay.} The left panel shows the CXR images, radiologist
    report, and image metadata, while the right panel shows the associated
    ED stay, triage, vital-sign, and medication records. The
    \textit{Findings} and \textit{Impression} sections are used as the
    report-generation targets.}
    \label{fig:data_overview}
\end{figure}

We use a CXR report generation dataset constructed from two publicly available sources, MIMIC-CXR and MIMIC-IV-ED~\cite{johnson2019mimiccxr,johnson2023mimicived}. Chest X-ray exams from MIMIC-CXR are linked with the corresponding emergency department records from MIMIC-IV-ED to provide additional clinical context for report generation~\cite{nicolson2025impact}. Each sample consists of one or more CXR images from the same examination,
the reference radiology report, and four auxiliary patient-data sources:
triage, reconciled medicines, indication, and clinical history. We adopt
these sources following CXRMate-ED~\cite{nicolson2025impact}, whose
source-wise evaluation identified them as effective inputs and showed that
their combination without prior examinations achieved the strongest overall
performance among the evaluated patient-data configurations. We only include CXR exams performed within the time interval of an ED stay and remove events occurring after the imaging time~\cite{nicolson2025impact}. After preprocessing, the train/validation/test splits contain \textit{76,398/556/958 exams}, respectively. The task is to generate the two main sections of a radiology report: \textit{\textbf{Findings}}, which describes detailed observations from the images, and \textit{\textbf{Impression}}, which summarizes the key clinical conclusions. An example of a linked CXR examination and its corresponding ED record is
shown in Fig.~\ref{fig:data_overview}.

\subsection{Metrics}

We evaluate generated reports using two groups of metrics. Clinical metrics include RadGraph-F1 (RG)~\cite{delbrouck2022improving} for clinical entity and relation overlap, CheXbert-F1 (CX)~\cite{smit2020combining} for pathology-label agreement, CXR-BERT similarity (CB)~\cite{boecking2022making} for clinical semantic similarity, and GREEN (G)~\cite{ostmeier2024green} for clinically significant errors. Language generation metrics include BERTScore-F1 (BS)~\cite{zhang2020bertscore} for semantic similarity, ROUGE-L (R-L)~\cite{lin2004rouge} for longest-common-subsequence overlap, and BLEU-4 (B4)~\cite{papineni2002bleu} for n-gram overlap. We additionally report Absence of Repeated N-grams~\cite{nicolson2025impact} (ARN) to measure repetition avoidance, and the average prompt length to assess compression efficiency.


\subsection{Implementation Details}
Our implementation follows CXRMate-ED and uses the architecture shown in
Fig.~\ref{fig:main_method}. We implement ICGF-CXRMate with UniFormer~\cite{li2023uniformer} as the image encoder and a Llama-based decoder~\cite{touvron2023llama} as the language decoder. The model is trained in three stages, as described in Section~\ref{sec:training_objectives}. Stage 1 and Stage 2 are trained for 10 epochs, while Stage 3 is trained for 5 epochs. In Stage 2, we use CXR images together with clinical information, including triage, reconciled medicines, indication, and history; the image encoder is frozen to focus on learning patient-data-related components. In Stage 3, the reward weights are set to $\lambda_1=0.45$, $\lambda_2=0.45$, $\lambda_3=0.10$, $\alpha_f=0.75$, and $\alpha_i=0.25$. The model is optimized using AdamW with a learning rate of $5\times10^{-5}$ for Stage 1 and Stage 2, and $5\times10^{-6}$ for Stage 3. The proposed image-conditioned compressor uses $K=16$ learnable queries and eight-head cross-attention to produce fixed-length clinical context tokens, which are integrated with the visual representations through an eight-head gated cross-attention module. All experiments are conducted on an NVIDIA L40 (46GB) GPU.

\section{Main Results}

Table~\ref{tab:main_results} reports report-level performance on the
combined \textit{Findings} and \textit{Impression} sections. For
CXRMate-ED, incorporating auxiliary patient data in Stage~2 improves several
metrics over the image-only Stage~1 model, including CB from 53.85 to 56.86
and BLEU-4 from 4.26 to 5.14. However, direct concatenation increases the
average prompt length by 119.23 tokens, from 254.66 to 373.89, illustrating
the additional input cost of the complete patient-data sequence.

In Stage~2, ICGF-CXRMate outperforms CXRMate-ED across all seven clinical
and language-generation metrics. In particular, it improves RG from 24.00
to 28.86, CB from 56.86 to 62.80, BS from 21.47 to 25.82, and BLEU-4 from
5.14 to 7.96. Meanwhile, the average prompt length decreases from 373.89 to
288.37 tokens, corresponding to a reduction of 85.52 tokens, or 22.9\%,
with only a small decrease in ARN from 96.18 to 95.69. These results show
that the proposed compression and fusion mechanisms improve report quality
while substantially reducing the decoder input.

After self-critical sequence training in Stage~3, ICGF-CXRMate outperforms
CXRMate-ED on six of eight report-quality metrics while preserving the
shorter prompt. The largest improvements are observed in RG, GREEN,
ROUGE-L, and BLEU-4, whereas CX and ARN are slightly lower. This indicates
that the proposed architecture remains effective under sequence-level
optimization, although small trade-offs remain in pathology classification
agreement and repetition avoidance.

\begin{table}[t]
\caption{\textbf{Report-level performance on the combined
\textit{Findings} and \textit{Impression} sections.}
Best results within each training stage are shown in \textbf{bold}.}
\label{tab:main_results}
\centering
\scriptsize
\setlength{\tabcolsep}{2.5pt}
\renewcommand{\arraystretch}{1.08}

\resizebox{\textwidth}{!}{%
\begin{tabular}{lccccccccc}
\toprule
\textbf{Method}
& \textbf{RG$\uparrow$}
& \textbf{CX$\uparrow$}
& \textbf{CB$\uparrow$}
& \textbf{G$\uparrow$}
& \textbf{BS$\uparrow$}
& \textbf{R-L$\uparrow$}
& \textbf{B4$\uparrow$}
& \textbf{ARN$\uparrow$}
& \textbf{Prompt} \\
\midrule

\multicolumn{10}{c}{
\textbf{Stage 1: Images only}
} \\
\midrule

CXRMate-ED
& 23.46
& 27.16
& 53.85
& 31.44
& 20.25
& 24.44
& 4.26
& 95.00
& 254.66 \\

\midrule
\multicolumn{10}{c}{
\textbf{Stage 2: Images + triage, reconciled medicines, indication, and history}
} \\
\midrule

CXRMate-ED
& 24.00
& 27.32
& 56.86
& 32.75
& 21.47
& 25.46
& 5.14
& \textbf{96.18}
& 373.89 \\


\textbf{ICGF-CXRMate (Ours)}
& \textbf{28.86}
& \textbf{30.90}
& \textbf{62.80}
& \textbf{35.25}
& \textbf{25.82}
& \textbf{28.43}
& \textbf{7.96}
& 95.69
& 288.37 \\

\midrule
\multicolumn{10}{c}{
\textbf{Stage 3: Reinforcement learning from Stage 2}
} \\
\midrule

CXRMate-ED
& 30.12
& \textbf{32.61}
& 73.07
& 36.53
& 29.08
& 26.51
& 4.29
& \textbf{95.89}
& 373.89 \\


\textbf{ICGF-CXRMate (Ours)}
& \textbf{33.90}
& 32.34
& \textbf{74.12}
& \textbf{40.00}
& \textbf{32.07}
& \textbf{31.49}
& \textbf{8.02}
& 95.15
& 288.37 \\

\bottomrule
\end{tabular}%
}
\end{table}

\subsection{Additional Analysis}

\subsubsection{Component Ablation}

\begin{table}[!t]
\caption{\textbf{Component ablation on the combined
\textit{Findings} and \textit{Impression} sections in Stage 2.}
Best and second-best results are shown in \textbf{bold} and
\underline{underlined}, respectively.}
\label{tab:component_ablation}
\centering
\scriptsize
\setlength{\tabcolsep}{2.2pt}
\renewcommand{\arraystretch}{1.10}

\resizebox{\textwidth}{!}{%
\begin{tabular}{c|c|c|ccccccccc}
\toprule

\textbf{Img.-cond.}
& \textbf{Gated Fusion}
& \textbf{Direct $Z$}
& \textbf{RG$\uparrow$}
& \textbf{CX$\uparrow$}
& \textbf{CB$\uparrow$}
& \textbf{G$\uparrow$}
& \textbf{BS$\uparrow$}
& \textbf{R-L$\uparrow$}
& \textbf{B4$\uparrow$}
& \textbf{ARN$\uparrow$}
& \textbf{Prompt} \\

\midrule

\xmark
& \xmark
& \cmark
& \underline{28.39}
& 28.99
& \underline{60.75}
& 34.70
& 25.35
& \underline{28.15}
& \textbf{8.13}
& 95.54
& 288.37 \\

\cmark
& \xmark
& \cmark
& 24.58
& 26.39
& 55.82
& 32.99
& 21.97
& 25.81
& 5.19
& 95.09
& 288.37 \\

\xmark
& \cmark
& \cmark
& 28.35
& \underline{30.79}
& 59.41
& 34.28
& \underline{25.48}
& 28.10
& 7.74
& \textbf{95.86}
& 288.37 \\

\cmark
& \cmark
& \xmark
& 25.84
& 28.84
& 53.06
& \underline{34.94}
& 22.62
& 26.52
& 5.54
& 95.42
& 272.37 \\

\cmark
& \cmark
& \cmark
& \textbf{28.86}
& \textbf{30.90}
& \textbf{62.80}
& \textbf{35.25}
& \textbf{25.82}
& \textbf{28.43}
& \underline{7.96}
& \underline{95.69}
& 288.37 \\

\bottomrule
\end{tabular}%
}
\end{table}

\textit{Img.-cond.} indicates whether the queries are conditioned on the CXR representations before retrieving patient information. \textit{Gated Fusion} indicates whether $Z$ updates the visual tokens through gated cross-attention, while \textit{Direct $Z$} indicates whether $Z$ is also included as a separate token sequence in the decoder prompt. Therefore, disabling Direct $Z$ removes only the direct clinical-token pathway; $Z$ may still influence the decoder through the fused visual tokens. 
Table~\ref{tab:component_ablation} shows that the full model achieves the best overall performance, confirming the complementary roles of image conditioning and gated fusion. Removing Direct $Z$ shortens the prompt by 16 tokens but substantially reduces several metrics. This indicates that the direct clinical-token pathway complements the indirect pathway through the fused visual representations.

\subsubsection{Effect of the Number of Clinical Context Tokens}

\begin{table}[!t]
\caption{\textbf{Effect of the number of clinical context tokens $K$ on
the combined \textit{Findings} and \textit{Impression} sections in Stage 2.}
Prompt denotes the average number of input prompt tokens.
Best and second-best results are shown in \textbf{bold} and
\underline{underline}, respectively.}
\label{tab:k_ablation_full_report}
\centering
\small
\setlength{\tabcolsep}{2.8pt}
\renewcommand{\arraystretch}{1.08}

\resizebox{\textwidth}{!}{%
\begin{tabular}{lccccccccc}
\toprule
\textbf{No. Tokens}
& \textbf{RG$\uparrow$}
& \textbf{CX$\uparrow$}
& \textbf{CB$\uparrow$}
& \textbf{G$\uparrow$}
& \textbf{BS$\uparrow$}
& \textbf{R-L$\uparrow$}
& \textbf{B4$\uparrow$}
& \textbf{ARN$\uparrow$}
& \textbf{Prompt} \\
\midrule

$K=8$
& 25.56
& 26.40
& 52.93
& \textbf{35.53}
& 22.85
& 25.98
& 5.19
& \textbf{96.71}
& 280.37 \\

$K=16$
& \textbf{28.86}
& \textbf{30.90}
& \textbf{62.80}
& \underline{35.25}
& \textbf{25.82}
& \textbf{28.43}
& \underline{7.96}
& 95.69
& 288.37 \\

$K=32$
& 27.81
& 28.82
& 59.13
& 34.13
& 25.16
& 27.88
& 7.57
& \underline{95.87}
& 304.37 \\

$K=48$
& \underline{28.50}
& \underline{30.88}
& \underline{60.58}
& 34.26
& \underline{25.64}
& \underline{28.12}
& \textbf{8.13}
& 95.55
& 320.37 \\

\bottomrule
\end{tabular}%
}
\end{table}

Table~\ref{tab:k_ablation_full_report} shows that $K=16$ provides the best
  overall trade-off, achieving the highest scores on six metrics with a
  relatively compact prompt. Although $K=8$ yields the shortest prompt and the
  best GREEN and ARN scores, it performs worse on most other metrics. Increasing
  $K$ to 32 or 48 lengthens the prompt without consistent improvements, although
  $K=48$ obtains the highest BLEU-4 score. We therefore use $K=16$ in the final
  model.

\subsubsection{Qualitative Case Study}

\begin{figure}[t]
    \centering
    \includegraphics[width=\textwidth]{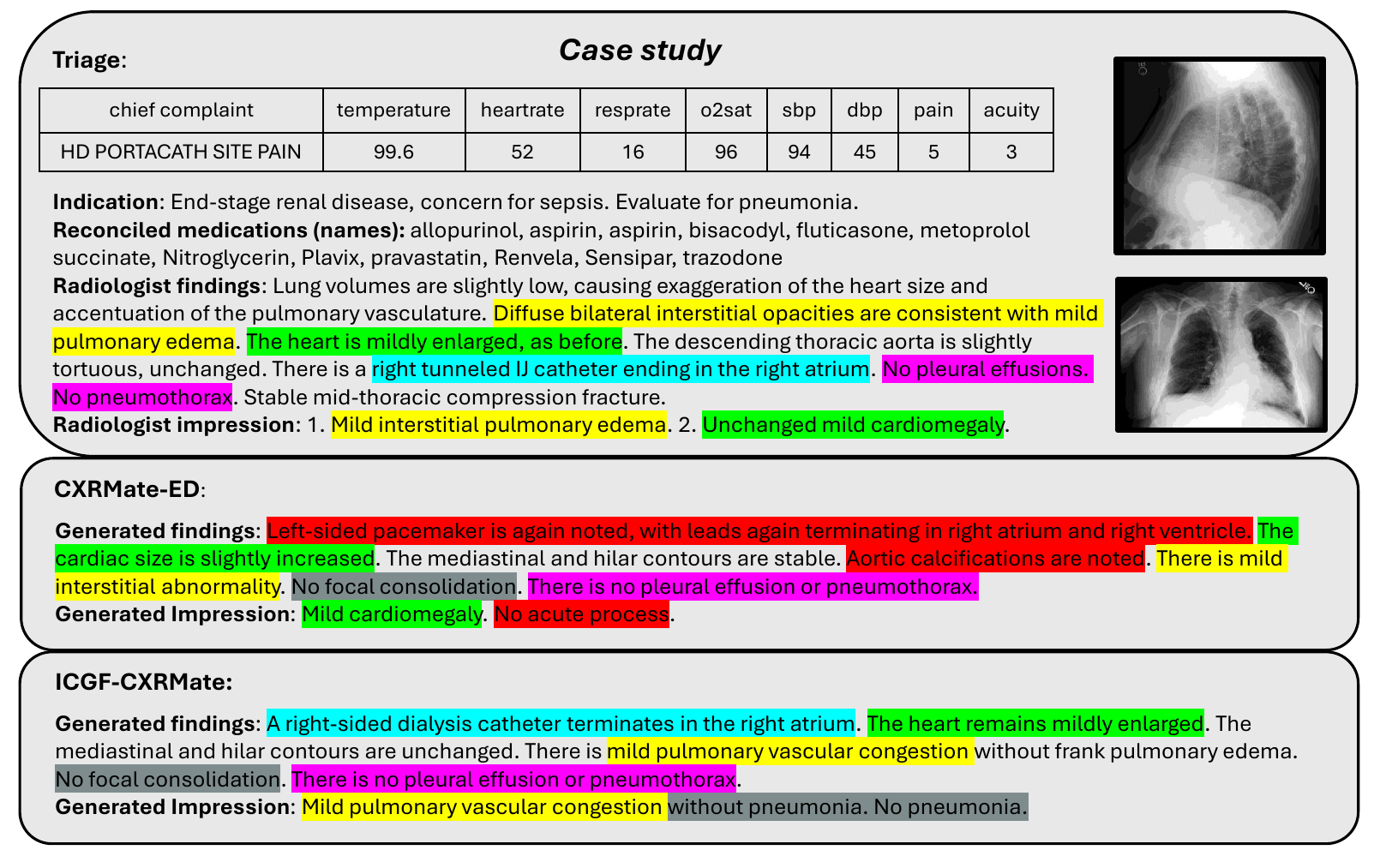}
    \caption{Case study comparing CXRMate-ED and ICGF-CXRMate ($K=16$).}
    \label{fig:example}
\end{figure}

As shown in Fig.~\ref{fig:example}, the patient presented with end-stage
renal disease and pain at the hemodialysis catheter site. CXRMate-ED
hallucinated a left-sided pacemaker and failed to capture the acute
interstitial abnormality. In contrast, ICGF-CXRMate correctly identified the
right-sided dialysis catheter terminating in the right atrium and preserved
the finding of mild cardiomegaly. It also detected mild pulmonary vascular
congestion, although this only partially matched the reference finding of
mild interstitial pulmonary edema. Overall, this case demonstrates that
image-conditioned clinical context improves examination-relevant device
identification and cardiopulmonary interpretation.

\section{Conclusion}
\label{sec:conclusion}

We introduced ICGF-CXRMate, a framework for incorporating heterogeneous
auxiliary patient data into chest X-ray report generation through
image-conditioned clinical context compression and gated cross-attention
fusion. The proposed compressor transforms variable-length patient
information into a fixed number of radiographically relevant clinical
tokens, while gated fusion selectively integrates this context into the
visual representations. Retaining both the compressed clinical tokens and
the fused visual tokens provides complementary direct and indirect clinical
pathways to the language decoder.

Experiments on linked MIMIC-CXR and MIMIC-IV-ED examinations demonstrate
that ICGF-CXRMate improves report quality while reducing the average prompt
length by 22.9\% compared with CXRMate-ED. The model also outperforms
CXRMate-ED on six of eight report-quality metrics after self-critical
sequence training. Ablation studies confirm the complementary roles of
image conditioning, gated fusion, and direct retention of the compressed
clinical context. The current evaluation is limited to the linked MIMIC
benchmark and does not include external validation. Future work will
investigate adaptive compression and evaluation on multi-institutional
datasets.

%
%
%
%

\bibliographystyle{splncs04}
\bibliography{sn-bibliography}
\end{document}